\documentclass[lettersize,journa,twoside]{IEEEtran}

\IEEEoverridecommandlockouts
\usepackage{cite}
\usepackage{amsmath,amssymb,amsfonts}
\usepackage{algorithmic}
\usepackage{graphicx}
\usepackage{subcaption}
\usepackage{mwe} 
\usepackage{textcomp}
\usepackage{xcolor}
\usepackage{lipsum}                     
\usepackage{xargs}                      
\usepackage[colorinlistoftodos,prependcaption,textsize=tiny]{todonotes}
\usepackage{array}
\usepackage{caption}
\usepackage{float}
\usepackage{multirow}
\usepackage{booktabs} 
\usepackage{tablefootnote}
\usepackage{threeparttable}
\usepackage{hyperref}
\usepackage{colortbl}
\def\BibTeX{{\rm B\kern-.05em{\sc i\kern-.025em b}\kern-.08em
    T\kern-.1667em\lower.7ex\hbox{E}\kern-.125emX}}

\usepackage{xspace}
\usepackage{hyperref}

\newcommand{\ML}{ML-MCTS\xspace}
\newcommand{\SP}{H=4s\xspace}
\newcommand{\Traj}{Traj. Opt.\xspace}
\newcommand{\MDM}{MPDM\xspace}
\newcommand{\POMDP}{POMDP\xspace}
\newcommand{\DP}{Decoupled\xspace}
\newcommand{\NOISTO}{W/O-CSTO\xspace}
\newcommand{\NOIS}{W/O-IS\xspace}
\newcommand{\NOUNC}{W/O-Unc.\xspace}

\newcommand{\eg}{e.g.}
\newcommand{\etc}{etc.}

\newcommand{\algname}{Hi-Drive\xspace}

\renewcommand{\eqref}[1]{(\ref{#1})}
\newcommand{\tabref}[1]{Table~\ref{#1}}

\newcommand{\figref}[1]{Fig.~\ref{#1}}

\newcommand{\egocar}{ego-vehicle\xspace}

\usepackage{etoolbox}
\newtoggle{final}
\definecolor{darkgreen}{rgb}{0 0.6 0}
\definecolor{darkblue}{rgb}{0 0 0.8}
\definecolor{darkred}{rgb}{0.8 0 0}
\definecolor{black}{rgb}{0 0 0}
\definecolor{lightgreen}{rgb}{0.529 0.784 0.333}
\newcommand{\xj}[1]{\iftoggle{final}{#1}{{\color{black} #1}}}
\newcommand{\cd}[1]{\iftoggle{final}{#1}{{\color{black} #1}}}

\usepackage{fontawesome}

\newcommand{\blackup}{\textcolor{black}{$\uparrow$}}
\newcommand{\blackdown}{\textcolor{black}{$\downarrow$}}

\newcommand{\ColSpaceA}{\hspace{0.4em}}

\makeatletter
\let\NAT@parse\undefined
\makeatother
\usepackage{hyperref}  

\begin{document}

\title{\bfseries
\algname: Hierarchical POMDP Planning for Safe Autonomous Driving in Diverse Urban Environments\\
}

\author{Xuanjin Jin, Chendong Zeng, Shengfa Zhu, Chunxiao Liu, and Panpan Cai
\thanks{Manuscript received: April 8, 2025; Revised: August 7, 2025; Accepted: September 24, 2025. This paper was recommended for publication by Editor Ashis Banerjee upon evaluation of the Associate Editor and Reviewers’ comments. This work was supported in part by the National Natural Science Foundation of China under Grant 62303304. \textit{(Corresponding Author: Panpan Cai.)}}
\thanks{Xuanjin Jin, Chendong Zeng and Panpan Cai are with Shanghai Jiao Tong University, Shanghai 201100, China, and also with Shanghai Innovation Institute, Shanghai 200030, China (e-mail: xuanjin.jin@sjtu.edu.cn; chendongzeng@sjtu.edu.cn; cai\_panpan@sjtu.edu.cn).}
\thanks{Shengfa Zhu and Chunxiao Liu are with SenseTime, Shanghai 200240, China. (e-mail: zhushengfa@senseauto.com; liuchunxiao@senseauto.com).}
\thanks{Digital Object Identifier (DOI): see top of this page.}
}







\maketitle
\markboth{IEEE ROBOTICS AND AUTOMATION LETTERS. PREPRINT VERSION. ACCEPTED October, 2025}%
{Jin \MakeLowercase{et al.}: Hi-Drive: Hierarchical POMDP Planning for Safe Autonomous Driving in Diverse Urban Environments}

\begin{abstract}
Uncertainties in dynamic road environments pose significant challenges for behavior and trajectory planning in autonomous driving. This paper introduces \algname, a hierarchical planning algorithm addressing uncertainties at both behavior and trajectory levels using a hierarchical Partially Observable Markov Decision Process (POMDP) formulation. \algname employs driver models to represent uncertain behavioral intentions of other vehicles and uses their parameters to infer hidden driving styles. By treating driver models as high-level decision-making actions, our approach effectively manages the exponential complexity inherent in POMDPs. To further enhance safety and robustness, \algname integrates a trajectory optimization based on importance sampling, refining trajectories using a comprehensive analysis of critical agents. Evaluations on real-world urban driving datasets demonstrate that \algname significantly outperforms state-of-the-art planning-based and learning-based methods across diverse urban driving situations in real-world benchmarks.

\begin{IEEEkeywords}
Autonomous driving, Uncertainty handling, Partially Observable Markov Decision Process (POMDP), Hierarchical Planning, Behavior and Trajectory Planning
\end{IEEEkeywords}

\end{abstract}


\section{Introduction}

Autonomous driving (AD) aims to improve transportation convenience by enabling vehicles to sense, plan and act autonomously. A critical component of AD is behavior and trajectory planning, which determines high-level driving behaviors (lane-following, lane-changing, turning, \etc) and low-level trajectories (specifying positions, speeds, and accelerations) of the autonomous vehicle (\textit{\egocar}). 
These decisions must be carefully made to ensure safety and efficiency.


A significant challenge in behavior and trajectory planning arises from uncertainty, particularly regarding unknown behavioral intentions and driving styles of other traffic participants (\textit{exo-agents}). For example, subtle actions of nearby vehicles, such as slowly drifting towards the lane boundary, can reflect diverse intentions. These could range from preparing to change lanes or trying to create space for another vehicle, to simply exhibiting reckless driving habits. Hedging against such uncertainties is essential for safe and robust planning.


Optimally addressing these uncertainties is computationally intractable in real-time settings, due to the exponential growth of computation with the number of traffic participants and the planning horizon, known respectively as the ``curse of dimensionality" and the ``curse of history" \cite{pineau_anytime_2006}. To mitigate complexity, existing approaches typically simplify uncertainty modeling by addressing only specific aspects, such as behavior-level uncertainty \cite{galceran_multipolicy_2017, schmerling_multimodal_2018,  ding_epsilon_2021, brechtel_probabilistic_2014, sunberg_improving_2022, gonzalez_human-like_2019, xia_interactive_2022, shu_autonomous_2021, bai_intention-aware_2015, danesh_leader_2023, naghshvar_risk-averse_2018}, trajectory-level uncertainty \cite{speidel_graph-based_2021, xu_motion_2014, lutzow_density_2023,  li_real-time_2015}, or by employing maximum-likelihood assumptions \cite{li_efficient_2022, hoel_combining_2019, lenz_tactical_2016, albrecht_interpretable_2021}. However, these simplifications can compromise safety and robustness, especially in scenarios involving complex interactions among vehicles.


\xj{In this work, we propose Hi-Drive, a hierarchical POMDP planner that, for the first time, enables joint reasoning over uncertainties in both high-level behavioral intentions and low-level driving styles for general urban driving.}
We treat the behavioral intentions and driving styles of exo-agents as latent states, represented using driver models and their free parameters, respectively. We develop a hierarchical Bayesian filter to track a hierarchy of probability distributions over these hidden states (\textit{hierarchal belief}) over time, enabling structured and efficient reasoning about complex and uncertain interactions.

To address the computational complexity inherent to POMDP planning, \algname also uses high-level driver models to represent actions for the \egocar, thereby shortening the effective planning horizon. At the lower level, driving trajectories are generated by simulating these driver models, allowing joint evaluation of decisions across both behavior and trajectory levels. Additionally, to enhance trajectory robustness, a subsequent trajectory optimization step employs importance sampling, re-generating and re-evaluating trajectories, with a focus on critical agents.



Experimental evaluation shows the effectiveness of \algname for safe and robust long-term planning in  urban environments. Evaluations on the large-scale nuPlan benchmark \cite{caesar_nuplan_2022} show that \algname achieves competitive performance compared to learning-based approaches, despite requiring no training. Further analysis on an additional real-world dataset, featuring long-term tasks, shows substantial improvements in safety and task completion compared to existing planning-based methods. We attribute \algname's advantage to hierarchical and long-term reasoning and the application of importance sampling. 
Furthermore, \algname can perform real-time planning in complex driving scenes, even with limited computational resources. 
It also allows transparent interpretation of its reasoning, as shown in the supplementary video.

In summary, our main contributions include:
\begin{itemize} 
\xj{\item A hierarchical POMDP planner that synergistically and efficiently handles uncertainties at both behavior and trajectory levels. }
\item Integration of POMDP planning with an importance-sampling-based trajectory optimizer to enhance trajectory safety and robustness. 
\item This work is the first to apply POMDP planning to general urban scenes (instead of pre-selected scenes such as multi-lane roads) and achieve state-of-the-art performance in large-scale benchmarks. 
\end{itemize}


\section{Related Work}

\begin{figure*}[t]
\centering{\includegraphics[width=17.2cm]{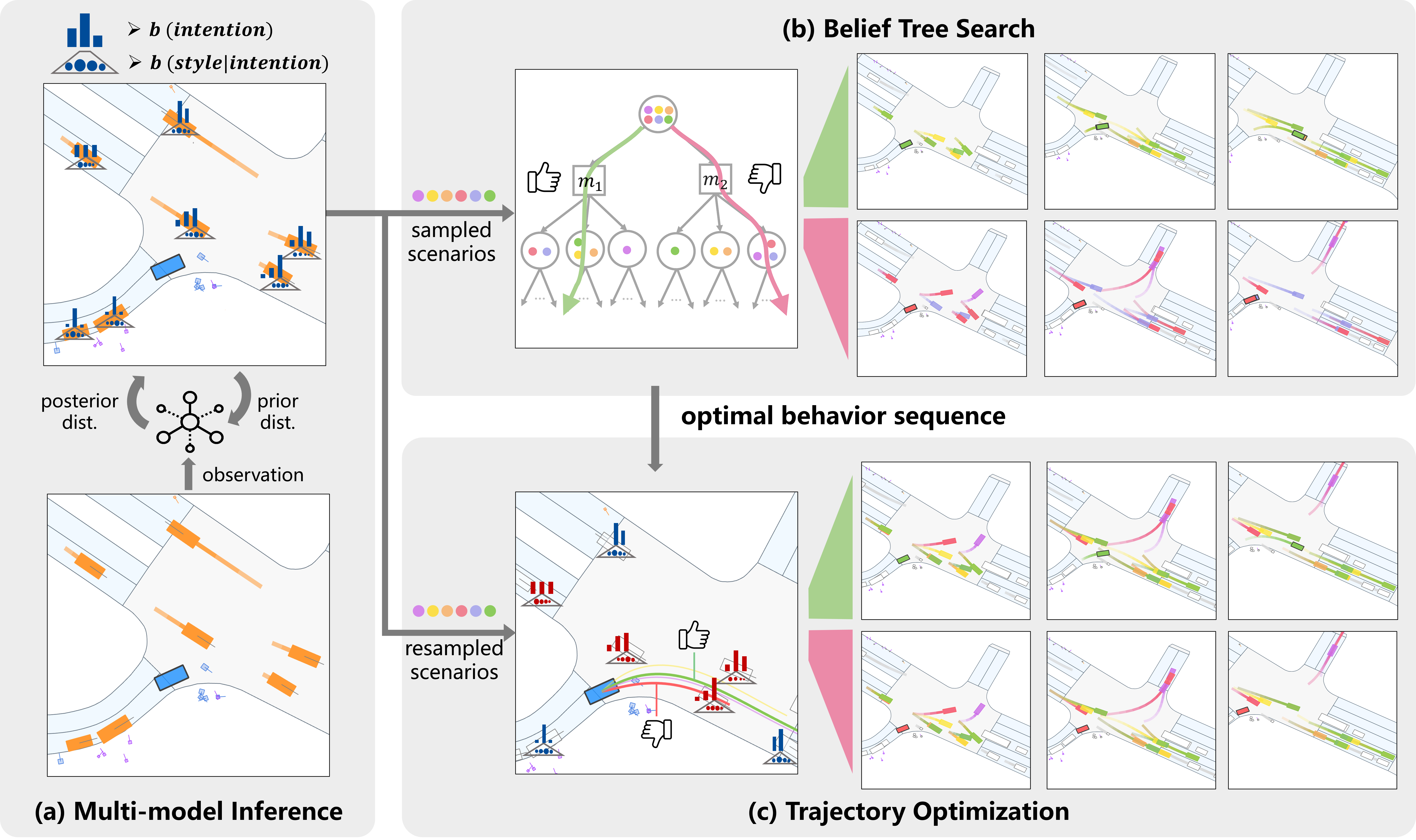}}
\caption{\cd{Overview of \algname: (a) The multi-model inference tracks beliefs about exo-agents' behavioral intentions and driving styles based on real-time observations. (b) The belief tree search determines the \egocar’s optimal policy over high-level behaviors. (c) Trajectory optimization resamples representative scenarios using importance sampling, cross-evaluates low-level trajectories generated by those scenarios, and selects the one with the \textcolor{lightgreen}{best} driving performance.
See detailed explanations of the driving scene visualization in Section \ref{interpretation}.}
}
\vspace*{-0.5cm}
\label{fig::overview}
\end{figure*}

Safe and robust autonomous driving requires hedging against uncertainties at both behavior and trajectory levels. High-level uncertainties involve predicting others' routes, maneuvers, or interaction decisions, while low-level uncertainties correspond to variations in acceleration, steering, \etc \ However, optimally tackling both levels of uncertainty leads to intractable computational complexity, thus, it requires simplification to achieve real-time performance.
Planning under uncertainty is often simplified to make it computationally tractable. A straightforward approach is to plan for the maximum-likelihood behaviors of others, commonly using Monte-Carlo Tree Search (MCTS)~\cite{li_efficient_2022, hoel_combining_2019, albrecht_interpretable_2021, lenz_tactical_2016}. While efficient, this may pose risks in scenes with low-probability events.

\xj{Other methods address the behavior-level uncertainty or trajectory-level uncertainty in isolation. This separation, coupled with prior focus on specific environments like highways, compromises safety in complex and general urban scenarios where both intentions and styles interact.} For instance, to handle trajectory uncertainty, planners use algorithms like RRT* or numerical optimization, but these are often decoupled from a high-level behavioral planner, which prohibits synergistic behaviors like changing lanes to avoid an unpredictable driver~\cite{lutzow_density_2023, speidel_graph-based_2021, xu_motion_2014, li_real-time_2015}. To handle behavior uncertainty, Multipolicy Decision-Making (MPDM) evaluates a set of candidate policies for the ego-vehicle~\cite{galceran_multipolicy_2017, schmerling_multimodal_2018, ding_epsilon_2021}. However, its open-loop nature ignores the benefits of future information, which can lead to overly conservative actions.

To perform closed-loop planning under uncertainty, past works formulate the driving problem as POMDPs, treating the unknown behaviors of others as hidden states, then apply POMDP solvers such as QMDP\cite{naghshvar_risk-averse_2018}, POMCP \cite{gonzalez_human-like_2019, sunberg_improving_2022, xia_interactive_2022}, DESPOT \cite{bai_intention-aware_2015, danesh_leader_2023}, and ABT \cite{shu_autonomous_2021} to compute the \egocar's behavior policy conditioned on future observations. 
Some of these works perform short-term planning by searching over primitive actions, such as longitudinal accelerations and lateral velocities \cite{brechtel_probabilistic_2014, sunberg_improving_2022, xia_interactive_2022, shu_autonomous_2021, bai_intention-aware_2015, danesh_leader_2023,bey_handling_2021}, while others consider high-level behaviors such as lane keeping and changing for long-term planning \cite{gonzalez_human-like_2019, naghshvar_risk-averse_2018}. 
However, existing POMDP methods are typically limited to specific scenes, such as multi-lane roads. Moreover, they only address behavior-level uncertainty and neglect trajectory-level uncertainty for computational simplicity. This is insufficient given that different drivers with the same intention may exhibit vastly different trajectories—e.g., an aggressive driver might cut into a lane abruptly, while a conservative driver would do so more cautiously. In this work, we extend the POMDP approach to tackle urban environments and achieve state-of-the-art results on a large-scale benchmark, nuPlan \cite{caesar_nuplan_2022}. 
We also use a hierarchical POMDP framework to address uncertainty at both behavior and trajectory levels simultaneously, and complement it with a trajectory optimizer based on importance sampling to further enhance driving safety and robustness within limited planning time.

\xj{Our work also advances on prior hierarchical and factored POMDPs by introducing a synergistic framework that, unlike sequential switching of granularity \cite{saleem2024pomdp,zhao2021hierarchical}, jointly reasons about high-level intentions and low-level styles. We also use an agent-centric factorization of hidden states that, in contrast to state-region, observability, or objective-based decompositions \cite{jain2018efficient,ong2009pomdps,sheng2024safe,carr2021safe}, ensures scalability with the number of agents.
}

\section{Overview} 

In this paper, we propose \algname, a hierarchical planner (\figref{fig::overview}) addressing uncertainties at both behavior and trajectory levels.
Our approach is grounded in a Partially Observable Markov Decision Process (POMDP) framework, modeling behavioral intentions and driving styles of exo-agents as hidden states, and treating the behavioral intention of the \egocar as actions. 

\algname consists of three interconnected modules:

\textit{Multi-model Inference} (\figref{fig::overview}a):
Behavioral intentions of exo-agents are represented by distinct driver models, while driving styles are characterized by their parameters. We employ hierarchical Bayesian inference—combining exact filtering at the intention level and particle filtering at the style and state levels—to dynamically update and track beliefs about exo-agents' behaviors based on observations.

\textit{Belief Tree Search} (\figref{fig::overview}b):  
Conditioned on the updated beliefs, we perform an online belief tree search to identify the optimal high-level behavior for the \egocar. The algorithm samples \emph{driving scenarios} (see Section~\ref{sec::scenario}) from the current belief, then recursively explores possible actions and resulting observations to estimate their future outcomes and uncertainties. An optimal behavior sequence is thus determined from this belief-space planning process.

\textit{Trajectory Optimization} (\figref{fig::overview}c):  
Following the selected optimal behavior sequence, we apply importance sampling to resample representative driving scenarios, emphasizing the analysis of critical agents. These sampled scenarios are then used to refine low-level driving trajectories for the \egocar. Finally, a cross-scenario evaluation selects the most robust and effective trajectory for execution.



\section{Hierarchical POMDP Modeling}
Our approach is centered on a hierarchical POMDP formulation, using high-level driver models as its building blocks.

\subsection{Driver Models}
We consider two primary driving behaviors: lane-following (\textit{LF}) and lane-changing (\textit{LC}). Each behavior is represented by a driver model:

\begin{itemize} 
    \item \textit{Lane-following (LF)} guides vehicles along the centerlines of lanes and through lane connectors at intersections.
    \item \textit{Lane-changing (LC)} uses the MOBIL decision-making model \cite{kesting_general_2007}, which ensures sufficient gap and minimal interference with the target lane. \end{itemize}

Both behaviors employ pure pursuit \cite{coulter_implementation_1992} for lateral trajectory control and the Intelligent Driver Model (IDM) \cite{treiber_congested_2000} for longitudinal acceleration control. Driving style differs by behavior: for lane-following, style is represented by IDM’s desired speed (higher values indicate greater aggressiveness); for lane-changing, it is characterized by the look-ahead distance in pure pursuit (larger distances indicate more cautious driving).

Multiple \textit{LF} or \textit{LC} options can exist simultaneously, \eg, vehicles can select different outgoing lanes at intersections. To reduce computational complexity, we dynamically filter behaviors to include only legal actions consistent with the road network topology.


\subsection{POMDP Model}
The POMDP model is formulated as follows:



\subsubsection{State and Observation}
A state in our POMDP formulation includes:
\begin{itemize} 
    \item Behavioral intentions (driver models) of exo-agents: $\cd{m=\{m_i\}_{i\in I\mathrm{exo}}}$. 
    \item Driving styles (model parameters) of exo-agents: $\cd{\theta=\{\theta_i\}_{i\in I\mathrm{exo}}}$. 
    \item Physical states of all agents: $\cd{x=\{x_{\mathrm{ego}}\}\cup \{x_i\}_{i\in I{\mathrm{exo}}}}$, comprising positions, speeds, headings, \etc 
\end{itemize}

Observations $\cd{o=\{o_{\mathrm{ego}}\}\cup\{o_i\}_{i\in I{\mathrm{exo}}}}$ reflect noisy measurements of the agents' physical states.



\subsubsection{Action Space}
Actions for the \egocar are high-level behaviors defined also by the driver models (\textit{LF}s, \textit{LC}s). Formally, an action is $\cd{a = (m_{\mathrm{ego}}, \theta_{\mathrm{ego}})}$, where $\cd{m_{\mathrm{ego}}}$ represents the behavior to decide for the \egocar , and $\cd{\theta_{\mathrm{ego}}}$ is a pre-defined style value.


\subsubsection{State Transition and Observation Models}
The state transition function is deterministic and given by $\cd{x^t = G(x^{t-1}, m,\theta, m_{\mathrm{ego}}, \theta_{\mathrm{ego}}})$, where $G$ simulates one step forward using the specified driver models and style parameters.
The observation model assumes Gaussian noise around the true physical states: $p(o^t|x^t)=\mathcal{N}(o^t;x^t, \Sigma)$.


\subsubsection{Reward Function} \label{sec::reward_model}
The reward function evaluates key performance aspects:
\begin{itemize}
\item \textit{Safety}: Cubic penalty for collisions, scaled by speed.
\item \textit{Efficiency}: Linear penalty based on deviation from desired speed.
\item \textit{Task Completion}: Exponential penalty for deviation from the target lane.
\item \textit{Smoothness}: Constant penalty for each lane change.
\end{itemize}

\section{Belief Tracking with Multi-Model Inference} \label{sec::MULTI_MODEL}
This section presents multi-model inference for effectively tracking exo-agents' behavioral intentions and driving styles within the POMDP framework.

\subsection{Independence Assumptions}
Direct Bayesian inference over the high-dimensional hidden states of our POMDP, which include $N$ discrete intention variables and approximately $N\times M$ continuous style parameters (for $N$ exo-agents and $M$ intentions per agent), is computationally prohibitive. To achieve real-time performance, we employ two conditional independence assumptions:

\begin{itemize}
    \item Behavioral intentions of exo-agents are assumed independent during the inference window, allowing independent inference of the $N$ intention variables.
    \item Driving styles are assumed independent across different exo-agents and intentions, enabling independent inference of the $N\times M$ style parameters.
\end{itemize}

\xj{These assumptions yield a factored belief structure that significantly reduces inference complexity, making it linear with respect to the number of agents and intentions.} Note that the independence applies \textit{only} to latent intentions and style parameters. Contextual interactions between agents remain fully modeled through the resulting behaviors, \eg, a \textit{LC} behavior falls back to \textit{LF} when the gap in the target lane is insufficient. 

Besides, we assume intentions and styles are static within each planning cycle. Although dynamic intentions could be modeled using a learned transition function, our experiments suggest that, in practice, it has minimal benefit under computational constraints.


\subsection{Hierarchical Bayesian Filter}
We maintain the factorized belief using hierarchical Bayesian filtering, consisting of low-level particle filters and high-level exact filters.

\subsubsection{Low-Level Particle Filter} 
For each intention hypothesis $m_{i,k}$ of exo-agent $i$, we run a particle filter~\cite{djuric_particle_2003} to approximate the belief over driving style $\theta_i$ and physical state $x_i^t$:
\begin{equation}\nonumber
b^t(x_i^t, \theta_i \mid m_i = m_{i,k})
\end{equation}
Each particle filter consists of a \textit{prediction step}, propagating particles forward via deterministic transition, $\cd{x^t_i = G_i(x^{t-1}, m_i, \theta_i, m_{\mathrm{ego}}, \theta_{\mathrm{ego}})}$, and an \textit{update step} adjusting particle weights based on observation likelihoods, $p(o^t_i|x^t_i)=\mathcal{N}(o^t_i;x^t_i, \Sigma)$.


\subsubsection{High-Level Exact Filter} 
We apply exact Bayesian filtering at the intention level. The belief update for exo-agent $i$'s intention $m_i$ is computed as:
\begin{equation}\label{eqn::high-level-filter}\nonumber
b^t(m_i) = \eta\, p(o_i^t \mid m_i)\, b^{t-1}(m_i),
\end{equation}
where $\eta$ is a normalization constant. 

The challenge is to efficiently calculate the observation likelihood $p(o_i^t \mid m_i)$.  We derive:
\begin{equation}\label{eqn:m-likelihood}\nonumber
\small
p(o_i^t \mid m_i) = \sum_{x_i^{t-1}, \theta_i} p\left(o_i^t \mid G_i(x_i^{t-1}, m_i, \theta_i)\right) p(x_i^{t-1}, \theta_i \mid m_i).
\end{equation}
This equation suggests upward information propagation, integrating observation likelihoods of low-level particles at the current step, $p\left(o_i^t \mid G_i(x_i^{t-1}, m_i, \theta_i)\right)$, with particle weights from the last step, $p(x_i^{t-1}, \theta_i \mid m_i)$, to compute the high-level likelihood. Reuse of information enables efficient inference.



\begin{figure}[t]
\centering{\includegraphics[width=8cm]{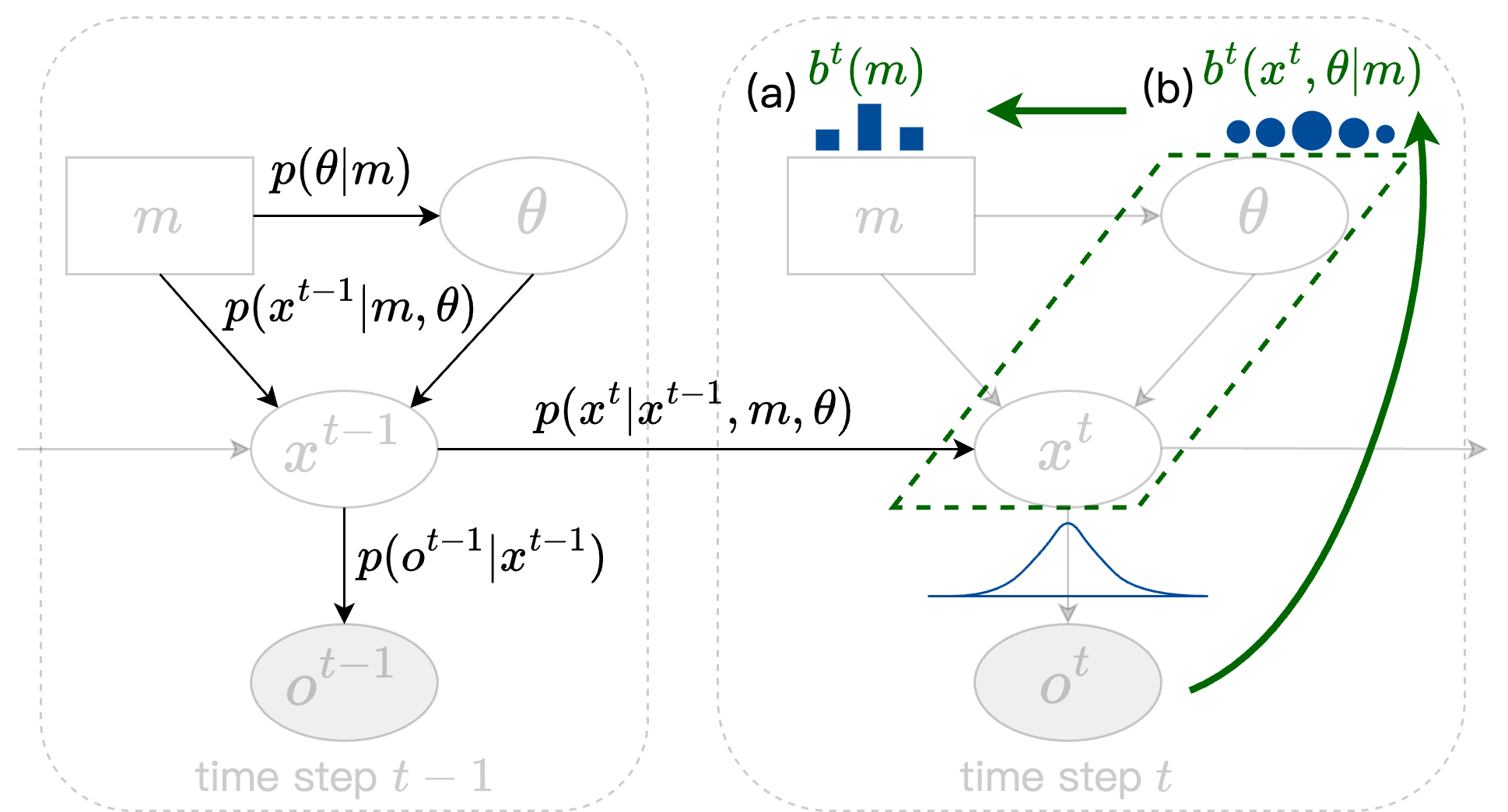}}
\caption{The hierarchical belief tracker. Green arrows represent the process of updating the belief. (a) The belief over behavioral intention $m$. (b) The belief over physical states $x^t$ and driving styles $\theta$.}
\label{fig::bayesian}
\vspace{-0.4cm}
\end{figure}

\section{Scenario-Based Belief Tree Search} \label{sec::behavior and trajectory planner}
Conditioned on the current belief, our belief tree search performs hierarchical behavior and trajectory planning, based on the POMDP formulation. The search algorithm is built upon DESPOT \cite{ye_despot_2017}, an online POMDP planner that guarantees asymptotic optimality, extended here to support hierarchical beliefs and varying-length actions.


\subsection{Sampling Driving Scenarios}
\label{sec::scenario}


The root belief node contains sampled \textit{driving scenarios}. Each scenario $\phi$ is generated by sampling behavioral intentions (driver models, $m$) from the high-level exact filter and driving styles (model parameters, $\theta$) and physical states ($x$) from low-level particle filters. These are combined with a sequence of random numbers $(\psi_t, \psi_{t+1}, \dots, \psi_{t+D})$ to determinize future actions and observations, where $D$ is the maximum search depth. Formally, a scenario is expressed as $\phi = (x, m, \theta, \psi_t, \psi_{t+1}, \dots, \psi_{t+D})$, succinctly encoding hypothetical dynamics of the driving scene using $N$ intention values, $N$ style values and $D$ random numbers. The belief tree is constructed based on $K$ scenarios to ensure fast computation, while maintaining asymptotic optimality.

\subsection{Abstract Action and Observation Branching}
Action branching occurs when expanding each belief node, representing the \egocar's selectable driving behaviors. 
\xj{The action space is dynamically constructed from the union of legal behaviors available to scenarios within that belief. Each action—such as a 2-second Lane-Follow (LF) or a 4-second Lane-Change (LC)—advances its corresponding scenarios using the forward simulation model $x^t = G(x^{t-1}, m, \theta, m_{\mathrm{ego}}, \theta_{\mathrm{ego}})$.} 
Each completed action creates a new action node containing the updated scenarios. It also carries a step reward calculated as the cumulative reward averaged over the encountered scenarios. 



Observation branching occurs under each action node, reflecting that \egocar's observations may differ under various scenarios. Observations are deterministically generated using the \textit{determinized} observation function $Z(o_t|x_t, \psi_t)$, by sampling Gaussian noise in $p(o_t|x_t)$ with $\psi_t$. Continuous observations are uniformly discretized, forming abstract observation branches. Each branch creates new a belief node containing a subset of scenarios inherited from the parent.

\xj{Abstract branching greatly reduced the search complexity, while the hierarchical belief structure maintained in belief nodes enables coupled evolution of behaviors and trajectories for all agents.}

\subsection{Iterative Tree Construction}
The belief tree iteratively expands via exploration paths. The exploration adopt DESPOT's heuristics, selecting the action branch with the highest upper bound and the observation branch with the highest Weighted Excess Uncertainty (WEU, see \cite{ye_despot_2017}). It expands encountered leaf nodes, and proceeds until a maximum look-ahead horizon (e.g., 9 seconds) is reached, or the WEU at the leaf node becomes negative. 
Heuristic values are computed for the terminal leaf node.
Afterwards, the search starts a \textit{value backup} process towards the root, updating the values of all nodes throughout the path, following the Bellman's principle. Upon reaching the root, a new exploration path starts.




\subsection{Extracting Optimal Behavior Sequence}
The belief tree search terminates when a planning time limit (e.g., 0.1 second) is reached. Unlike typical online planners that output only the immediate optimal action, we extract an optimal action sequence from the tree, to facilitate long-term trajectory planning. This is done by traversing the highest-value action branches and the most-likely observation branches. 

This procedure also generates driving trajectories for scenarios encountered under the path. However, these trajectories have not been reliably evaluated due to sparsity of the belief tree. It thus requires an additional trajectory optimization module for robust planning.




\begin{table*}[!t]
    \centering
    \caption{Comparison with Learning-based Models.}
    \renewcommand{\arraystretch}{1.5}
    \setlength{\extrarowheight}{-1pt}
    \setlength{\tabcolsep}{1.9em} 
    \begin{tabular}{llcccccc}
        \toprule
        \multirow{2}{*}{\textbf{Type}} & \multirow{2}{*}{\textbf{Planner}} & \multicolumn{2}{c}{\textbf{Val14}} & \multicolumn{2}{c}{\textbf{Test14-random}} & \multicolumn{2}{c}{\textbf{Test14-hard}} \\
        \cmidrule(lr){3-4} \cmidrule(lr){5-6} \cmidrule(lr){7-8}
        & & \textbf{R} & \textbf{NR} & \textbf{R} & \textbf{NR} & \textbf{R} & \textbf{NR} \\
        \midrule
        \textcolor{gray}{\textit{Expert}} & \textcolor{gray}{Log-replay} & \textcolor{gray}{80.32} & \textcolor{gray}{93.53} & \textcolor{gray}{75.86} & \textcolor{gray}{94.03} & \textcolor{gray}{68.80} & \textcolor{gray}{85.96} \\
        \midrule
        \multirow{2}{*}{Learning-based} 
        & PLUTO & 78.11 & 88.89 & 78.62 & 89.90  & 59.74 & 70.03 \\
        & Diffusion Planner & 82.80 & 89.87 & 82.93 & 89.19 & 69.22 & 75.99 \\
        \midrule
        \multirow{3}{*}{Hybrid} 
        & PDM-Hybrid & 92.11 & 92.77 & 91.28 & 90.10 & 76.07 & 65.99 \\
        & PLUTO w/ refine. & 76.88 & 92.88 & 90.29 & 92.23 & 76.88 & 80.08 \\
        & Diffusion Planner w/ refine. & \underline{92.90} & \textbf{94.26} & 91.75 & \textbf{94.80} & 82.00 & 78.87 \\
        \midrule
        \multirow{4}{*}{Model-based} 
        & \algname (Ours) & \textbf{93.15} & \underline{93.62} & \underline{92.31} & 93.71 & \textbf{83.18} & 81.41 \\
        & \algname-Dynamic (Ours) & \textbf{93.15} & 93.51 & \textbf{92.56} & \underline{93.72} & \underline{83.17} & \underline{81.88} \\
        & \xj{\algname-QCNet (Ours)} & \cd{92.84} & \cd{93.42} & \cd{91.80} & \cd{93.37} & \cd{82.69} & \cd{\textbf{81.91}}\\
        & \xj{\algname-Uniform (Ours)} & \cd{92.80} & \cd{92.60} & \cd{91.62} & \cd{92.08} & \cd{82.56} & \cd{79.98} \\
        \bottomrule
    \end{tabular}

   \label{table::comparision_baselines}
    \vspace{-0.2cm}
\end{table*}

\begin{table*}[t]
    \centering
    \caption{Comparison with Model-based Planning.}
    \renewcommand{\arraystretch}{1.5}
    \setlength{\extrarowheight}{-1pt}
    \setlength{\tabcolsep}{0.9em} 
    \begin{tabular}{ll@{\ColSpaceA}c@{\ColSpaceA}c@{\ColSpaceA}c@{\ColSpaceA}c@{\ColSpaceA}cccccccc}
        \toprule
        \multirow{3}{*}{\textbf{Type}} & \multirow{3}{*}{\textbf{Planner}} & \multicolumn{5}{c}{\textbf{Private Long-Term}} & \multicolumn{2}{c}{\textbf{Val14}} & \multicolumn{2}{c}{\textbf{Test14-random}} & \multicolumn{2}{c}{\textbf{Test14-hard}}  \\
        \cmidrule(lr){3-7} \cmidrule(lr){8-9} \cmidrule(lr){10-11} \cmidrule(lr){12-13}
        & & \multicolumn{5}{c}{\textbf{NR}} & \multicolumn{1}{c}{\textbf{R}} & \multicolumn{1}{c}{\textbf{NR}} & \multicolumn{1}{c}{\textbf{R}} & \multicolumn{1}{c}{\textbf{NR}} & \multicolumn{1}{c}{\textbf{R}} & \multicolumn{1}{c}{\textbf{NR}}\\
        \cmidrule(lr){3-7} \cmidrule(lr){8-8} \cmidrule(lr){9-9} \cmidrule(lr){10-10} \cmidrule(lr){11-11} \cmidrule(lr){12-12} \cmidrule(lr){13-13}
        & & Reward \blackup  & Coll.R. (\%)\blackdown & MGR (\%) \blackdown & TTG (\emph{s}) \blackdown  & Comfort \blackup & Score & Score & Score  & Score & Score& Score \\
        \midrule
        \multirow{2}{*}{\shortstack{w/o \\ unc.}}
        & \DP & -0.90 & \underline{1.70} & \textbf{0.00} & 42.74 & \underline{0.96} & 88.38 & 88.38 & 89.09 & 91.06 & 78.66 & 75.43 \\
        & \ML & \underline{-0.88} & 2.00 & \textbf{0.00} & 42.75 & \textbf{0.98} & \underline{93.01} & 92.91 & \underline{92.08} & 93.07 & 82.64 & \underline{81.30} \\
        \midrule
        \multirow{4}{*}{\shortstack{w/ \\ unc.}} 
        & \Traj & -2.43 & 10.00 & 0.90 & \textbf{36.46} & 0.92 & 72.63 & 72.99 & 70.32 & 71.95 & 65.00 & 60.75 \\
        & \MDM & -0.94 & 1.90 & \textbf{0.00} & 42.37 & 0.94 & 89.71 & 89.06 & 88.66 & 89.86 & 77.97 & 75.51 \\
        & \POMDP & -0.89 & 1.80 & \underline{0.10} & \underline{41.36}& 0.95 & 92.92 & \underline{93.20} & 91.80 & \underline{93.17} & \underline{82.94} & 80.70  \\
        & \algname (Ours) & \textbf{-0.77} & \textbf{0.00} & \textbf{0.00}  & 41.44 & \textbf{0.98} & \textbf{93.15} & \textbf{93.62} & \textbf{92.31} & \textbf{93.71} & \textbf{83.18} & \textbf{81.41} \\
        \bottomrule
    \end{tabular}

    \label{table::model_based_comparision_baselines}
    \vspace{-0.2cm}
\end{table*}

\section{Cross-Scenario Trajectory Optimization with Importance Sampling}
The cross-scenario trajectory optimization conducts a more detailed analysis at the trajectory level, thereby enhancing robustness to uncertainty. 

\textit{Scenario Resampling: }
We employ importance sampling (IS) to resample driving scenarios, focusing specifically on \textit{critical agents}, defined as those near the \egocar or whose predicted paths intersect with the planned trajectory. Intentions of critical agents are uniformly resampled to ensure extensive coverage, while intentions and styles of non-critical agents are sampled directly from the current belief.

\textit{Trajectory Regeneration: }
For each resampled scenario, we regenerate the driving trajectory of the \egocar by forward simulating the optimal behavior sequence determined by the belief tree search. 
Exo-agents are simulated simultaneously based on their sampled intentions and driving styles.

\textit{Cross Evaluation: }
Each candidate trajectory for the \egocar is cross-evaluated against all resampled scenarios. The one  most robust under uncertainty is selected. Specifically, for each cross-evaluation trajectory-scenario pair, we simulate forward the world state conditioned on the \egocar's candidate trajectory. The resulting state sequence is evaluated using the reward model (Section \ref{sec::reward_model}).
The optimal trajectory is selected based on maximum expected value under the IS distribution $q$:
\begin{equation} \label{eqn::traj_opt_is}
\begin{aligned}
    E[V(\tau(\phi))] &= \frac{1}{n_{\text{IS}}}\sum_{i=1}^{n_{\text{IS}}}\prod_{j = 1}^N \frac{b(m_{i,j})}{q(m_{i, j})}V_i(\tau(\phi)),  \\
    \tau(\phi)^* &= \arg\max E[V(\tau(\phi))],
\end{aligned}
\end{equation}
where \(n_{\text{IS}}\) denotes the number of resampled scenarios, \(N\) is the number of exo-agents, \(\phi\) is a sampled scenario, and \(\tau(\phi)\) is a candidate trajectory generated under scenario \(\phi\). The trajectory’s value is represented by \(V(\tau(\phi))\). The importance weights \(\frac{b(m_{i,j})}{q(m_{i,j})}\) correct for sampling bias, ensuring unbiased estimation and thus robust trajectory selection.


\section{Experiments} \label{sec::experiments}
We compare \algname with both learning-based and planning-based methods on a wide variety of real-world driving scenes, including those from the nuPlan benchmark and a private dataset. 
Results show that \algname generalized well across diverse urban environments, improving driving safety and ensuring successful task completion. Compared to learning-based methods, although \algname requires no training, it achieved state-of-the-art performance in reactive settings and competitive results in non-reactive settings on nuPlan. 
Compared to existing planners, \algname is the only planner that achieved zero collisions and 100\% task success.
Ablation studies emphasize the significant contributions of hierarchical POMDP planning and cross-scenario trajectory optimization.
In the accompany\cd{ing} video, we also demonstrate how \algname's reasoning process can be transparently interpolated.



\begin{table*}[!t]
    \centering
    \caption{Ablation study results}
    \renewcommand{\arraystretch}{1.5}
    \setlength{\extrarowheight}{-1pt}
    \setlength{\tabcolsep}{1.25em} 
    \begin{tabular}{lc@{\ColSpaceA}c@{\ColSpaceA}c@{\ColSpaceA}c@{\ColSpaceA}ccccccc}
        \toprule
        \multirow{2}{*}{\textbf{Planner}}& \multicolumn{5}{c}{\textbf{Private Long-Term}} & \multicolumn{2}{c}{\textbf{Val14}} & \multicolumn{2}{c}{\textbf{Test14-random}} & \multicolumn{2}{c}{\textbf{Test14-hard}} \\
        \cmidrule(lr){2-6} \cmidrule(lr){7-8} \cmidrule(lr){9-10} \cmidrule(lr){11-12} 
        & \multicolumn{5}{c}{\textbf{NR}} & \multicolumn{1}{c}{\textbf{R}} & \multicolumn{1}{c}{\textbf{NR}} & \multicolumn{1}{c}{\textbf{R}} & \multicolumn{1}{c}{\textbf{NR}} & \multicolumn{1}{c}{\textbf{R}} & \multicolumn{1}{c}{\textbf{NR}}\\
        \cmidrule(lr){2-6} \cmidrule(lr){7-7} \cmidrule(lr){8-8}  \cmidrule(lr){9-9} \cmidrule(lr){10-10} \cmidrule(lr){11-11} \cmidrule(lr){12-12}
        & Reward \blackup  & Coll.R. (\%)\blackdown & MGR (\%) \blackdown & TTG (\emph{s}) \blackdown  & Comfort \blackup & Score & Score & Score & Score & Score & Score  \\
        \midrule
        \NOUNC & -0.88 & 2.00 & \textbf{0.00} & 42.75 & \textbf{0.98} & \underline{93.01} & 92.91 & 92.08 & 93.07 & 82.64 & \underline{81.30} \\
        \NOISTO & -0.89 & 1.80 & \underline{0.10} & \textbf{41.36}& \underline{0.95} & 92.92 & 93.20 & 91.80 & 93.17 & \underline{82.94} & 80.70 \\
        \NOIS & \underline{-0.79} & \underline{1.00} & \underline{0.10} & \underline{41.39} & \textbf{0.98} & 92.92 & \underline{93.23} & 92.09 & 93.42 & 82.88 & 81.26 \\
        \SP & -0.81 & \textbf{0.00} & 1.10 & 43.09  & \textbf{0.98} & 92.98 & 93.02 & \underline{92.20} & \underline{93.65} & 82.66 & 80.43 \\
        H=9s (Full) & \textbf{-0.77} & \textbf{0.00} & \textbf{0.00} & 41.44 & \textbf{0.98} & \textbf{93.15} & \textbf{93.62} & \textbf{92.31} & \textbf{93.71}  & \textbf{83.18} & \textbf{81.41} \\
        \bottomrule
    \end{tabular}
    \label{table::ablation_study}
    \hspace*{-0.05\linewidth}
    \parbox{0.9\linewidth}{
        \vspace{0.15cm}
        \small * \emph{\textbf{Bold} indicates best; \underline{underscored} indicates second-best.}
    }
    \vspace{-0.6cm}

\end{table*}

\subsection{Comparison with Learning-Based Models}
We first compare \algname to state-of-the-art learning-based methods on the nuPlan benchmark \cite{caesar_nuplan_2022}. The evaluation covers three datasets: \textit{Val14} \cite{Dauner2023CORL} (1,118 regular scenes), \textit{Test14-random} \cite{cheng2024rethinking} (268 random scenes), and \textit{Test14-hard} \cite{cheng2024rethinking} (272 complex scenes), with increasing difficulty. Each scenario lasts 15 seconds. We evaluate performance in both reactive and non-reactive settings using the official nuPlan metric, where higher scores indicate better performance.

Baselines include learning-based approaches—PLUTO \cite{cheng_pluto_2024} (contrastive imitation learning) and Diffusion Planner \cite{zheng2025diffusionbased} (diffusion models)—as well as hybrid methods that refine learned trajectories using optimization, such as \textit{PDM-Hybrid} \cite{Dauner2023CORL} (winner of the 2023 nuPlan challenge), PLUTO w. refine, and Diffusion Planner w. refine.

Results in \tabref{table::comparision_baselines} demonstrate that \algname consistently outperforms learning-based and hybrid methods in reactive scenarios, and achieves competitive results in non-reactive scenarios. Notably, in challenging scenarios (\textit{Test14-hard}), \algname significantly exceeds the performance of the best learning baseline. Note that \algname's performance is achieved without training.

\xj{
We also evaluated several variants of \algname: \textit{\algname-Uniform} maintains a fixed uniform belief over others' intentions, which results in overly conservative behaviors, showing the benefits of belief tracking; \textit{\algname-Dynamic} considers the intention dynamics of exo-agents by sampling future intentions from a prior distribution. Its similar performance shows that dynamics of future intentions play an insignificant role for online planning; 
lastly, \textit{\algname-QCNet} integrates trajectories predicted by a neural network (QCNet \cite{zhou2023query}), as an additional modality, to cover unstructured behaviors. Its similar performance shows that our selected driver model suite are indeed sufficient to model versatile behaviors.}

\subsection{Comparison with Model-Based Planning}
We further evaluate \algname against model-based planners. In addition to nuPlan, we introduce the \textit{Private Long-term} dataset, featuring 100 real-world scenes with 40-second episodes. 

To reveal detailed planning performance, we introduce new metrics including the overall performance represented by the average cumulative reward (\textit{reward}), safety performance represented by the collision rate among test episodes (\textit{Coll.R.}), task completion performance characterized by the rate of missing the goal lane segment (\textit{MGR}), driving efficiency denoted by the average time to reach goal (\textit{TTG}), and a comfort score related to lateral acceleration (\textit{Comfort}).

The compared planners include: \textit{\DP}, which follows a given global route with IDM, \textit{\ML} \cite{li_efficient_2022, hoel_combining_2019, albrecht_interpretable_2021, lenz_tactical_2016}, which performs MCTS with maximum-likelihood scenarios, \textit{\Traj} \cite{ li_real-time_2015}, which performs trajectory optimization under uncertainty, \textit{\MDM} \cite{ galceran_multipolicy_2017, schmerling_multimodal_2018,  ding_epsilon_2021}, representing multipolicy decision-making, and \textit{\POMDP} \cite{gonzalez_human-like_2019, sunberg_improving_2022}, representing belief tree search planning without trajectory optimization.

\tabref{table::model_based_comparision_baselines} shows that \algname outperforms all planners across both benchmarks. On nuPlan, which requires adapting to diverse urban environments, \algname consistently ranks highest in reactive and non-reactive settings. On \textit{Private Long-term}, which requires long-term planning and precise goal reaching, \algname achieves the safest driving and the highest task success.

Detailed analyses on \textit{Private Long-term} highlight \algname's advantages. Unlike planners without uncertainty modeling (\DP and \ML), \algname proactively mitigates risks, effectively reducing the collision rate to zero. Compared to planners with uncertainty modeling (\Traj, \MDM, \POMDP), \algname effectively integrates behavior and trajectory planning, achieving the highest cumulative reward as well as a notable 0\% miss-goal rate, while ensuring safety. 

\subsection{Ablation Study}
In this section, we conduct ablation studies to evaluate key components of \algname, by comparing with 4 variants: \NOUNC, that uses a single maximum-likelihood scenario for planning; \NOISTO, which omits cross-scenario trajectory optimization; \NOIS, which omits importance sampling in trajectory optimization;
and \SP, with shortened planning horizon from 9s to 4s.
Results in \tabref{table::ablation_study} illustrate the contributions of each component. \xj{Removing POMDP planning (\NOUNC) increases collision from 0\% to 2\%, due to passive and late response to risks; Removing cross-scenario trajectory optimization (\NOISTO) also increases collision by 1.8\%, due to insufficient granularity in trajectory analysis.
Removing importance sampling (\NOIS) in trajectory optimization reduces safety and goal-reaching, due to a lack of focus on rare and critical events.} Lastly, shortening the planning horizon (\SP) increases missed goals (1.1\%), highlighting the necessity of long-term planning.

\subsection{Computation Time}\label{planning_time}

\xj{\algname achieves real-time performance when running in a single CPU thread, handling over 20 exo-agents on average. The planning time meets the typical 10 Hz requirement for AD systems, measuring 94.3 ms on an Intel i7-8700K, 91.6 ms on a Xeon Gold 6330, and 112.4 ms on a Xeon Gold 6240C CPU. For the i7-8700K test, time is mainly spent on belief tree search (45.0 ms), followed by trajectory optimization (21.1 ms) and belief tracking (13.6 ms).}

\subsection{Interpretable Reasoning}\label{interpretation}
\figref{fig::overview}(b) and (c) illustrate the visual interpretation of the reasoning process of \algname in a scene of a right turn. A reduced number of driving scenarios are visualized for a clear presentation. Vehicles in each scenario are denoted in a unique color. 
In belief tree search (Fig.~\ref{fig::overview}b), a suboptimal action sequence (red) suggested turning into the rightmost lane, getting the \egocar stuck in the traffic flow. In contrast, the optimal action sequence (green) suggests turning into the left lane, which is less crowded, thus enables efficient progress. 
During trajectory optimization (Fig.~\ref{fig::overview}c), scenarios are resampled from an importance sampling distribution emphasizing critical agents (highlighted in red). Trajectories generated from the optimal action sequence are cross-evaluated with scenarios, selecting the green trajectory, which integrates seamlessly into traffic, and abandoned the red trajectory, which made little progress.
More detailed visualization of the reasoning process is included in the accompanying video.




\section{Conclusion}
This paper presents \algname, a hierarchical POMDP-based planner designed to handle uncertainties in behavior and trajectory planning for urban autonomous driving. \algname explicitly models uncertainties in other vehicles' behavioral intentions and driving styles, allowing effective belief tracking and robust decision-making. Our method combines hierarchical behavior planning with importance sampling-based trajectory optimization, achieving safe, efficient, and comfortable driving in diverse and complex urban environments. Experiments on real-world driving datasets validate that \algname outperforms both learning-based and model-based planning approaches, demonstrating strong safety and robustness.

\xj{
\algname still has the following limitations, inducing important directions of future work.
First, the current set of driver models we adopted were designed or trained for urban environments with HD-maps available. Future work could extend this to map-less environments by adding map-prediction modules \cite{xiong2023neural} or training new prediction models with map-less data. Second, we used a hand-crafted importance sampling distribution, which could be insufficient when handling very complex and crowded interactions; we plan to extend this to learned importance sampling distributions, for which \cite{danesh_leader_2023} offers a potential approach.
Furthermore, our current planner can be integrated with heuristic learning \cite{Lee-RSS-19} and macro-action learning \cite{Lee-RSS-21} to further enhance planning efficiency and performance. Parallelization can also be a significant aid to real-time performance, for example, through GPU parallelization \cite{cai2021hyp} or SIMD parallelization via instruction sets \cite{thomason2024motions}.
}

\section*{Acknowledgment}
Following the RAS policy, the use of generative AI is strictly limited to improve self-written texts to enhance readability. None of the presented methods and results (figures, equations, numbers, etc.) are generated by AI.

\end{document}